\documentclass[11pt]{article}

\usepackage[final]{acl}

\usepackage{times}
\usepackage{latexsym}

\usepackage[T1]{fontenc}

\usepackage[utf8]{inputenc}

\usepackage{microtype}

\usepackage{inconsolata}

\usepackage{graphicx}

\usepackage{multirow}
\usepackage{booktabs}
\usepackage{graphicx}
\usepackage{subcaption} 
\usepackage{amsmath}
\usepackage{amsfonts}
\usepackage{enumitem}
\usepackage{CJKutf8}
\usepackage{colortbl}
\usepackage{xcolor}
\usepackage[table]{xcolor} 

\usepackage{CJKutf8} 

\newcommand{\diff}[1]{\textcolor{black}{#1}}

\title{Data-efficient Targeted Token-level Preference Optimization\\
for LLM-based Text-to-Speech}

\author{
 \textbf{Rikuto Kotoge\textsuperscript{1,2}},
 \textbf{Yuichi Sasaki\textsuperscript{1,3}}
\\
\\
 \textsuperscript{1}SpiralAI Inc.,
 \textsuperscript{2}The University of Osaka,
 \textsuperscript{3}Shizuoka University
\\
 \small{
   \textbf{Correspondence:} \href{mailto:r_kotoge@go-spiral.ai}{r\_kotoge@go-spiral.ai}
 }
}

\begin{document}
\maketitle
\begin{abstract}
Aligning text-to-speech (TTS) system outputs with human feedback through preference optimization has been shown to effectively improve the robustness and naturalness of LLM-based TTS models.
Current approaches primarily require paired desirable and undesirable samples at the utterance level. However, such pairs are often limited in TTS output data, and utterance-level formulation prevents fine-grained token-level optimization needed for accurate pronunciation alignment.
In this study, we propose TKTO that eliminates the need for paired data, enabling a more data-efficient training paradigm, and directly targets token-level units, automatically providing fine-grained alignment signals without token-level annotations.
TKTO improves the challenging Japanese TTS accuracy by 39\% and reduces CER by 54\%, \diff{leveraging $6\times$ more training data and} assigning $12.8 \times$ stronger reward to targeted tokens. 
\end{abstract}

\section{Introduction}
\label{sec:intro}
\label{sec:intro}

\begin{figure}[t]
  \centering
    \centering
    \includegraphics[width=0.6\linewidth]{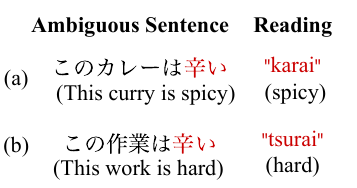}
  \caption{Examples of ambiguity in Japanese.
Although (a) and (b) contain the same word, its meaning and reading differ depending on the context.}
  \label{fig:example}
\end{figure}

Recent advances in neural network technology have enabled high-fidelity text-to-speech (TTS) synthesis models \cite{f5,melle,cm-tts}. They typically convert input text into a phoneme sequence using a grapheme-to-phoneme (G2P) converter \cite{openjtalk}, followed by generative models from the phoneme sequence \cite{wang2025maskgct,nishimurahall}.
However, G2P is generally based on morphological analysis, and thus may fail to generate correct pronunciations in ambiguous languages like Japanese, where the reading and meaning of words can change depending on context (Figure \ref{fig:example}).

G2P-free large language model (LLM)-based TTS methods \citep{wang2025spark,cosyvoice2} have shown great promise to generate context-aware pronunciations directly from raw text without G2P by leveraging large-scale natural language pretraining.
Recent work \citep{zhang-etal-2025-advancing-zero,tian2025preference,zhang2024speechalign} has further applied Direct Preference Optimization (DPO) \cite{dpo} to improve intelligibility, speaker similarity, and overall naturalness by enlarging the preference gap between pairwise samples.

Despite these advances, two fundamental challenges remain.
\textbf{(i) Necessity of paired data:} DPO-based methods require paired desirable and undesirable outputs for the same utterance, but TTS systems often produce one-sided results, where many samples are consistently desirable or undesirable. This causes significant data inefficiency and wastes costly human feedback, limiting the scalability of DPO-based preference alignment.
\textbf{(ii) Sample-level optimization:} Pronunciation generation is essentially a character- or token-level task, while preference alignment is conducted with  utterance-level labels. This mismatch forces the model to optimize at the data-sample level rather than directly at the pronunciation unit level, leading to suboptimal learning signals and limiting the effectiveness of alignment.

\begin{figure*}[t]
  \centering
    \centering
    \includegraphics[width=\linewidth]{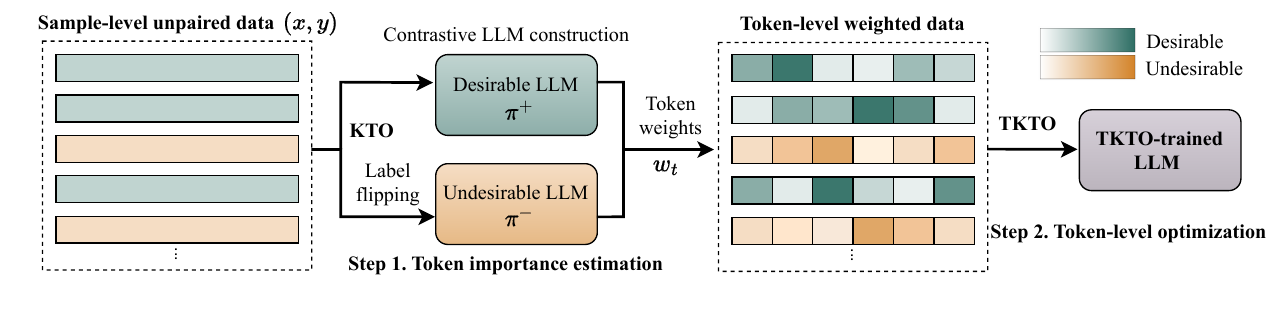}
  \caption{Overview of our TKTO framework. Step 1: we estimate token-level importance weights, constructing two contrastive LLMs. Step 2: we optimize token-level preferences.}
  \label{fig:ov}
  \vspace{-10pt}
\end{figure*}

To address these challenges, we propose a novel preference optimization framework, \textbf{Token-level Kahneman-Tversky Optimization (TKTO)} that constructs contrastive LLMs to estimate token-level weights and optimizes token-level preferences grounded in Kahneman-Tversky’s prospect theory \cite{kto}.
Our contributions are summarized in three key dimensions:
\begin{itemize}[nosep,leftmargin=*,topsep=0pt,partopsep=0pt,parsep=0pt,itemsep=0pt]
    \item \textbf{Problem formulation:} we pioneer the challenging task of ambiguous pronunciation as a token-level preference optimization problem.
    \item \textbf{Novel methodology:} we propose TKTO that (i) eliminates the need for paired data, enabling a more data-efficient training paradigm, and (ii) directly targets token-level units, automatically providing fine-grained alignment signals.
    \item \textbf{Wide evaluation:} we demonstrate that TKTO reduces character error rate (CER) by 54\% and improves the accuracy of Japanese pronunciation by 39 \%. 
    \diff{By leveraging unpaired data, TKTO can utilize $6\times$ more training data than DPO, leading to higher data efficiency.}
    It also automatically assigns 12.8× stronger rewards to targeted tokens.
\end{itemize}

\section{LLM-based Text-to-Speech}

We consider a text-to-speech (TTS) task formulated as conditional generation from input text $x$ to output speech token sequence $y = (y_1, \dots, y_T)$.
Let $\pi_\theta(y \mid x)$ denote a LLM decoder that autoregressively predicts acoustic tokens conditioned on textual input and previously generated tokens:
\begin{equation}
    \pi_\theta(y \mid x) = \prod_{t=1}^{T} \pi_\theta(y_t \mid x, y_{<t}).
\end{equation}
The output token sequence $y$ is transformed into speech audio via a neural vocoder.
Our goal is to train the LLM decoder $\pi_\theta$ to learn \emph{token-level preferences} from user feedback or preference data.

\begin{table*}[t]
  \centering
  \resizebox{0.85\linewidth}{!}{%
  \begin{tabular}{ll|ccc|ccc}
    \toprule
    \multirow{2}{*}{Model} & \multirow{2}{*}{PO Data} & \multicolumn{3}{c|}{Female} & \multicolumn{3}{c}{Male} \\
                           &                          & Acc ↑ & CER ↓ & Bad ↓ & Acc ↑ & CER ↓ & Bad ↓ \\
    \midrule    gpt-4o-mini-tts   & -        & 0.900 & 0.109 & 0.079    & 0.939 & 0.111 & 0.062 \\
    gemini-2.5-flash-preview-tts   & -        & 0.776 & 0.140 & 0.105    & 0.769 & 0.134 & 0.091 \\
    gemini-2.5-pro-preview-tts   & -        & 0.871 & 0.127 & 0.094    & 0.885 & 0.119 & 0.073 \\
    F5-TTS  \citep{f5}           & -        & 0.498 & 0.173 & 0.189 & 0.500 & 0.177 & 0.183 \\
    F5-TTS with G2P  \citep{openjtalk}     & -        & 0.500 & 0.136 & 0.100 & 0.500 & 0.146 & 0.107 \\

\midrule
    Base model \citep{cosyvoice2}        & -        & 0.683 & 0.128 & 0.090 & 0.668 & 0.138 & 0.095 \\
    Supervised Fine-Tuning (SFT)               & Desirable& 0.674 & 0.119 & 0.076 & 0.654 & 0.130 & 0.084 \\
    DPO \citep{tian2025preference}              & Paired     & 0.706 & 0.120 & 0.076 & 0.693 & 0.130 & 0.082 \\
    KTO \citep{kto}              & Paired     & 0.654 & \underline{0.066} & \underline{0.028} & 0.651 & \underline{0.074} & 0.030 \\
    KTO \citep{kto}               & Unpaired   & \underline{0.933} & 0.079 & 0.030 & \underline{0.952} & 0.087 & 0.032 \\
    \rowcolor{gray!15} TKTO (ours)   & Paired   & 0.681 & \textbf{0.059} & \textbf{0.025} & 0.701 & \textbf{0.066} & \underline{0.029} \\
    \rowcolor{gray!15} TKTO (ours)   & Unpaired   & \textbf{0.949} & \underline{0.075} & 0.029 & \textbf{0.958} & 0.085 & \textbf{0.027} \\
    \bottomrule
  \end{tabular}
  }
  \caption{Accuracy (Acc), Character Error Rate (CER), and bad case ratio (Bad) across different models. The best and second-best results are highlighted in \textbf{bold} and \underline{underline}, respectively.}
  \label{tab:main_table}
  
\end{table*}

\section{Proposed Method}
\label{sec:method}
We propose a two-step approach to optimize token-level preferences from unpaired data (Figure~\ref{fig:ov}). 
First, 
we quantify how informative each token is for preference learning.
These weights are then used to guide our TKTO objective.

\subsection{Targeted Token Weight Estimation}
We first construct two contrastive language models, $\pi^{+}$ and $\pi^{-}$, to capture token-level preferences. 
Then, we use the log-ratio between these two models to estimate token-level importance weights.

\paragraph{KTO-based Contrastive LLM Construction.} 
We propose a KTO-based approach to construct contrastive LLMs. 
KTO \cite{kto} uses unpaired data, enabling more flexible training.
\diff{Details are provided in Appendix~\ref{sec:kto}.
We build two contrastive LLMs, $\pi^{+}$ and $\pi^{-}$, 
which do not share parameters.} 
\diff{
We train $\pi^{+}$ using the original desirable/undesirable labels,
and obtain $\pi^{-}$ by training on the same data
with the labels swapped, i.e., treating desirable samples as undesirable
and vice versa.
}

\paragraph{Token-level Importance Sampling.}
We estimate the token's weight $w_t$ \cite{liu2025tisdpo} as:
\begin{equation}
   w_t = \exp(\mu \cdot \text{clamp}(\log\frac{\pi^{+}(y_t \mid x, y_{<t})}{\pi^{-}(y_t \mid x, y_{<t})}, L, U)),
   \label{eq:w_t_estimate}
\end{equation}
where $\log\frac{\pi^{+}(y_t \mid x, y_{<t})}{\pi^{-}(y_t \mid x, y_{<t})}$ estimates the token's reward \citep{rafailov2024from}. 
\diff{$L < U$ are lower and upper clamping bounds to improve optimization stability. $\mu$ is a scaling factor that controls the strength of reward weighting.}
For desirable samples, we set $\mu>0$; for undesirable ones, we set $\mu<0$.

\subsection{Token-level KTO}

We extend KTO \cite{kto} to the \emph{token level}, proposing Token-level KTO (TKTO) to learn finer-grained token-level signals.

\paragraph{Token-level Reward and Reference.}
We define the reward \diff{$r_{\theta,t}(x, y)$} for each token $y_t$ as its log-ratio against a reference policy $\pi_\text{ref}$:
\begin{align}
&r_{\theta,t}(x, y)
    = \log \frac{\pi_\theta\!\left(y_t \mid x, y_{<t}\right)}
                         {\pi_\text{ref}\!\left(y_t \mid x, y_{<t}\right)} ,
    \\
    &z_{0,t}
    = \mathrm{KL}\!\Big(
        \pi_{\theta}\!\left(\cdot \mid x, y_{<t}\right)
        \,\Big\|\, 
        \pi_\text{ref}\!\left(\cdot \mid x, y_{<t}\right)
      \Big) ,
\end{align}
where $z_{0,t}$ is a fixed reference baseline estimated across a microbatch (no gradients propagated).

\paragraph{Token-level Value Function.}
We define each token’s value $v_t$ using a logistic-shaped function:
\begin{align}
v_t(x, y) &=
    \begin{cases}
        \lambda_D\,\sigma\!\big(\beta \,(r_{\theta,t}(x,y) - z_{0,t})\big)
        \\ \quad \quad \quad \text{if } y \sim y_\text{desirable}\mid x, \\[4pt]
        \lambda_U\,\sigma\!\big(\beta \,(z_{0,t} - r_{\theta,t}(x,y))\big)
        \\ \quad \quad \quad \text{if } y \sim y_\text{undesirable}\mid x,
    \end{cases}
\end{align}
where $\sigma(\cdot)$ is the sigmoid function, $\beta$ controls curvature, and $\lambda_D$, $\lambda_U$ adjust aversion weights.

\paragraph{Objective Function.}
The TKTO loss is a weighted sum of token-level values, with importance weights $w_t$:
\vspace{-10pt}
\begin{align}
    L_{\text{TKTO}}
    \;=\;
    \mathbb{E}_{(x,y)}\!\left[
      -\sum_{t=1}^{|y|}
      w_t  \cdot  v_t(x,y) 
    \right].
    \label{eq:tkto_loss}
\end{align}

\section{Experiments}
\label{sec:exp}

\subsection{Experimental Setup}
\paragraph{Text Dataset.}
To evaluate not only CER but also more challenging cases of ambiguous Japanese pronunciation, we created a Japanese dataset of 5,000 sentences containing the word 
\begin{CJK}{UTF8}{min} "辛い" \end{CJK}
using GPT-5 \cite{openai_gpt5_2025}. 
The word  can be pronounced either as karai (spicy) or tsurai (hard), depending on the context; we ensured that each pronunciation appears in half of the samples.

\paragraph{Speech Dataset.}
For each text sentence, we generate five speech samples for both female and male Japanese speakers \cite{hi-fi-captain} using a TTS model, with training data containing 23 hours of speech for each speaker. Among them, the sample with the correct pronunciation and the lowest CER is selected as a desirable sample. Conversely, the sample with the incorrect pronunciation and the highest CER is selected as an undesirable sample.  
The CER is computed using the whisper-v3-large \cite{whisper}.

\paragraph{Baselines.}
We use CosyVoice2 (0.5B) \cite{cosyvoice2} fine-tuned on 20K hours of Japanese speech data as our base model and apply our TKTO. 
The following baselines are considered:

\begin{itemize}[nosep,leftmargin=*,topsep=0pt,partopsep=0pt,parsep=0pt,itemsep=0pt]
    \item \emph{Preference Optimization (PO) methods:}  
    DPO uses 1.5K paired desirable–undesirable samples;
KTO uses 9K unpaired desirable or undesirable samples;
SFT uses 6K desirable samples.

    \item \emph{TTS models:}  
    Flow matching-based F5-TTS ~\cite{f5} trained on the 20K hours, w/ and w/o Japanese G2P \cite{openjtalk}.  

    \item \emph{Reference industry models:}  
    gpt-4o-mini-tts \cite{openai_gpt4o_mini_tts_2025} (\textit{Coral}: female, \textit{Ash}: male) and gemini-2.5-tts-preview~\cite{google_gemini2_5_preview_tts_2025} (\textit{Zephyr}: female, \textit{Puck}: male).
\end{itemize}

\paragraph{Objective Metrics:}
\diff{Accuracy measures the proportion of samples in which the generated pronunciation matches the correct reading for a target ambiguous word (e.g., 
\begin{CJK}{UTF8}{min} "辛い" \end{CJK}
).
}
We also use CER to evaluate overall robustness 
and the bad case ratio (Bad), where samples with CER exceed 0.3.

\paragraph{Subjective Metrics: }
We employ the naturalness mean opinion score (NMOS) to evaluate the naturalness. point scale, where 1 indicates very unnatural and 5 indicates completely natural. 
In addition, we conduct an ABX test.
The instructions for the subjective evaluation are as follows.
Three Japanese native human annotators listened to the speech samples and rated their naturalness as NMOS on a 5-point scale, where 1 indicates very unnatural and 5 indicates completely natural. In addition, we conducted an ABX test in which the participant listened to two generated speech samples from different models but based on the same input and then chose the one that sounds more natural; if the samples are too similar to distinguish, they are instructed to indicate a tie. 
The annotator was recruited via a crowdsourcing platform and received appropriate compensation for their work.

\subsection{Objective Evaluation Results}

\begin{figure}[t]
  \centering
  \begin{subfigure}{0.25\textwidth}
    \centering
    \includegraphics[width=\linewidth]{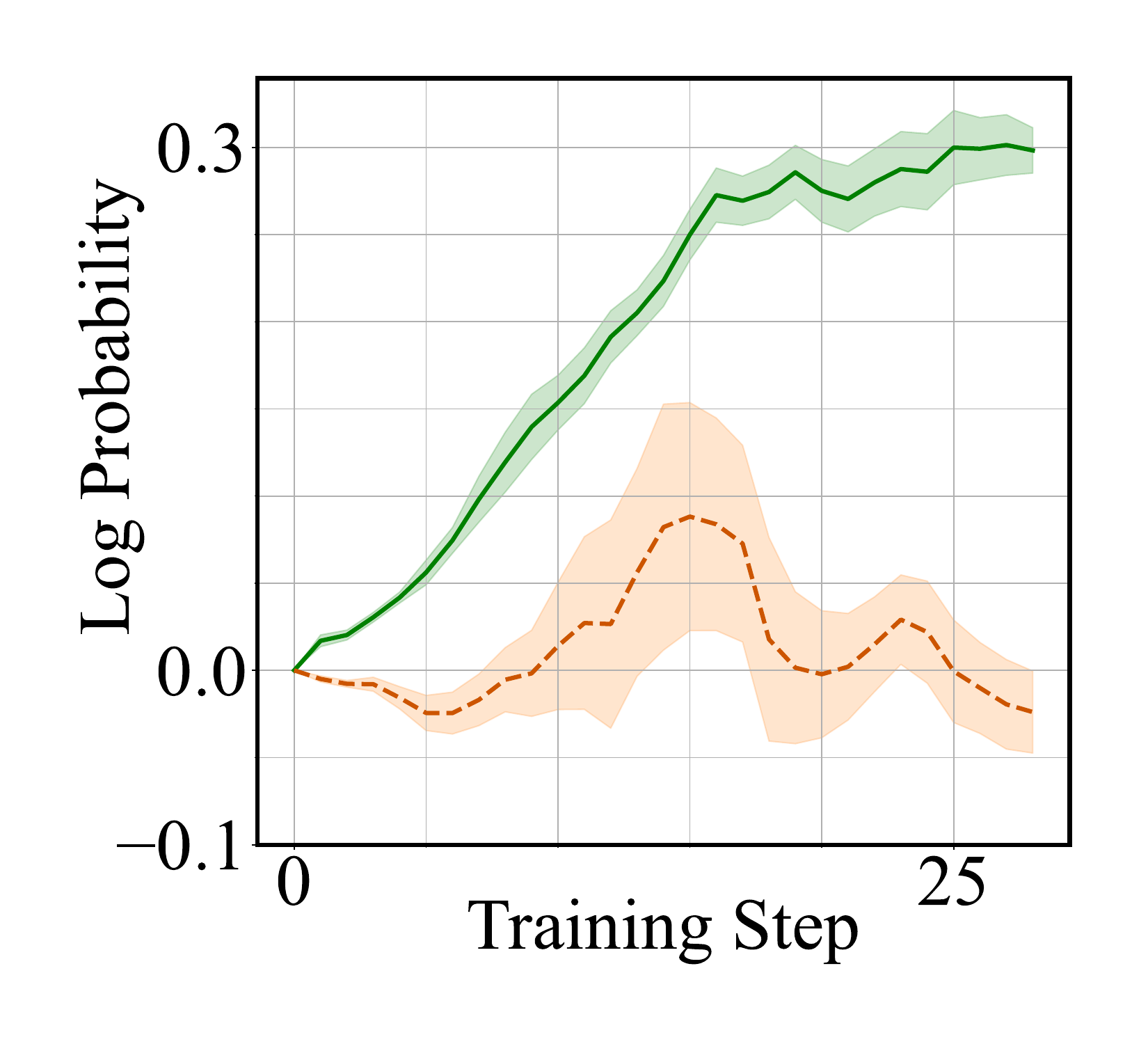}
    \caption{TKTO (Ours)}
    \label{fig:left}
  \end{subfigure}%
  \begin{subfigure}{0.25\textwidth}
    \centering
    \includegraphics[width=\linewidth]{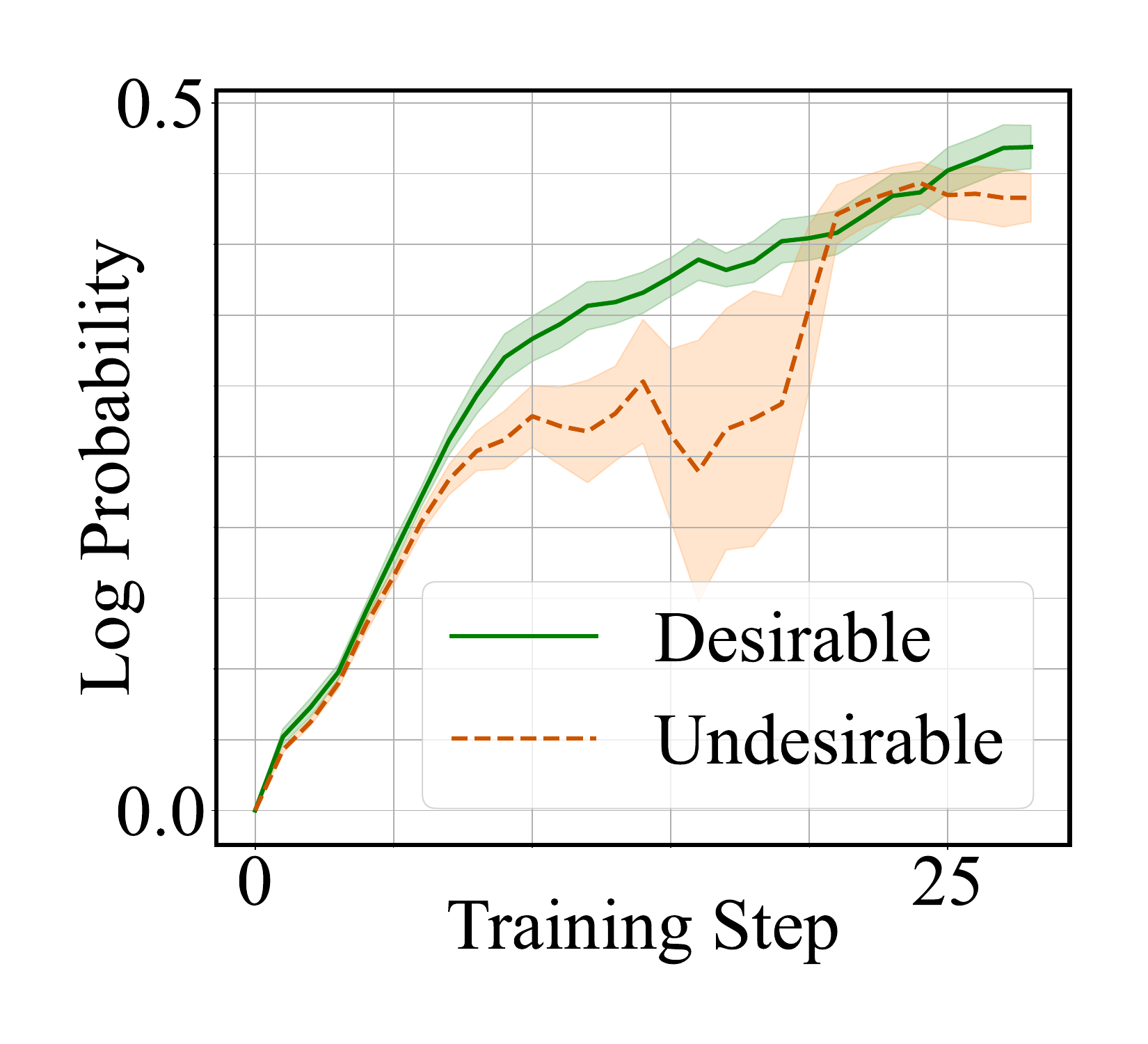}
    \caption{SFT}
    \label{fig:right}
  \end{subfigure}
  \caption{Average log-likelihood for desirable and undesirable tokens during training.  TKTO effectively increases only that of desirable tokens.}
  \label{fig:log-likelihood}
\end{figure}

\paragraph{Main Results on Japanese.}
Table~\ref{tab:main_table} presents the objective evaluation results.
Our TKTO model achieves the highest Japanese TTS accuracy and the lowest CER and bad case ratio.
The non-LLM baseline F5-TTS shows extremely low accuracy regardless of whether it uses G2P or not, highlighting the importance of LLMs to consider contextual information.
\diff{By leveraging unpaired data, we can utilize up to $6\times$ more training samples than paired methods.}
This leads to a significant improvement in accuracy (0.668→0.958), surpassing even strong industry models.
\diff{Paired training relies on a very narrow and biased subset that represents only rare borderline cases where both correct and incorrect pronunciations appear for the same text. This limits the learning signal and reduces generalization.}
Compared with KTO, our model achieves improvements across all metrics, demonstrating the effectiveness of token-level targeted optimization.

\paragraph{Training Dynamics.}
Figure~\ref{fig:log-likelihood} illustrates the changes in the average log-likelihood of desirable and undesirable tokens.
As training progresses, TKTO (left) effectively increases only that of desirable tokens, whereas SFT (right) tends to increase the log-likelihood of undesirable tokens.

\begin{figure}[t]
  \centering
    \centering
    \includegraphics[width=\linewidth]{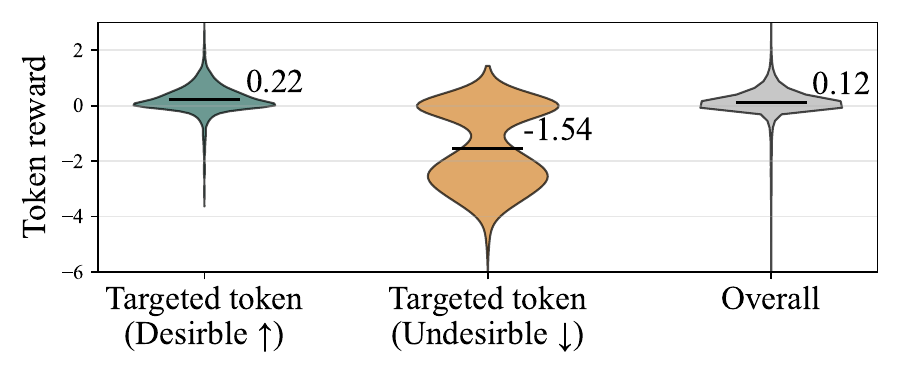}
  \caption{Token reward analysis. Desirable tokens have a higher reward, while undesirable tokens have a much lower reward, encouraging both weights to increase.}
  \label{fig:token_reward}
\end{figure}

\paragraph{Token Weight Analysis.}
Figure~\ref{fig:token_reward} illustrates the token rewards $\log\frac{\pi^{+}(y_t \mid x, y^{<t})}{\pi^{-}(y_t \mid x, y^{<t})}$ for desirable and undesirable tokens for the target character
\begin{CJK}{UTF8}{min} "辛" \end{CJK}
and the overall average.
The desirable tokens have a higher reward (0.22) compared to the overall mean (0.12), while the undesirable tokens show a much lower value (-1.54), indicating that targeted tokens are automatically assigned larger weights.
For undesirable samples, two peaks are observed when a clear difference between $\pi^{+}$ and $\pi^{-}$ emerges; the reward drops sharply, whereas when no significant difference exists.

\begin{figure}[t]
  \centering
    \centering
    \includegraphics[width=\linewidth]{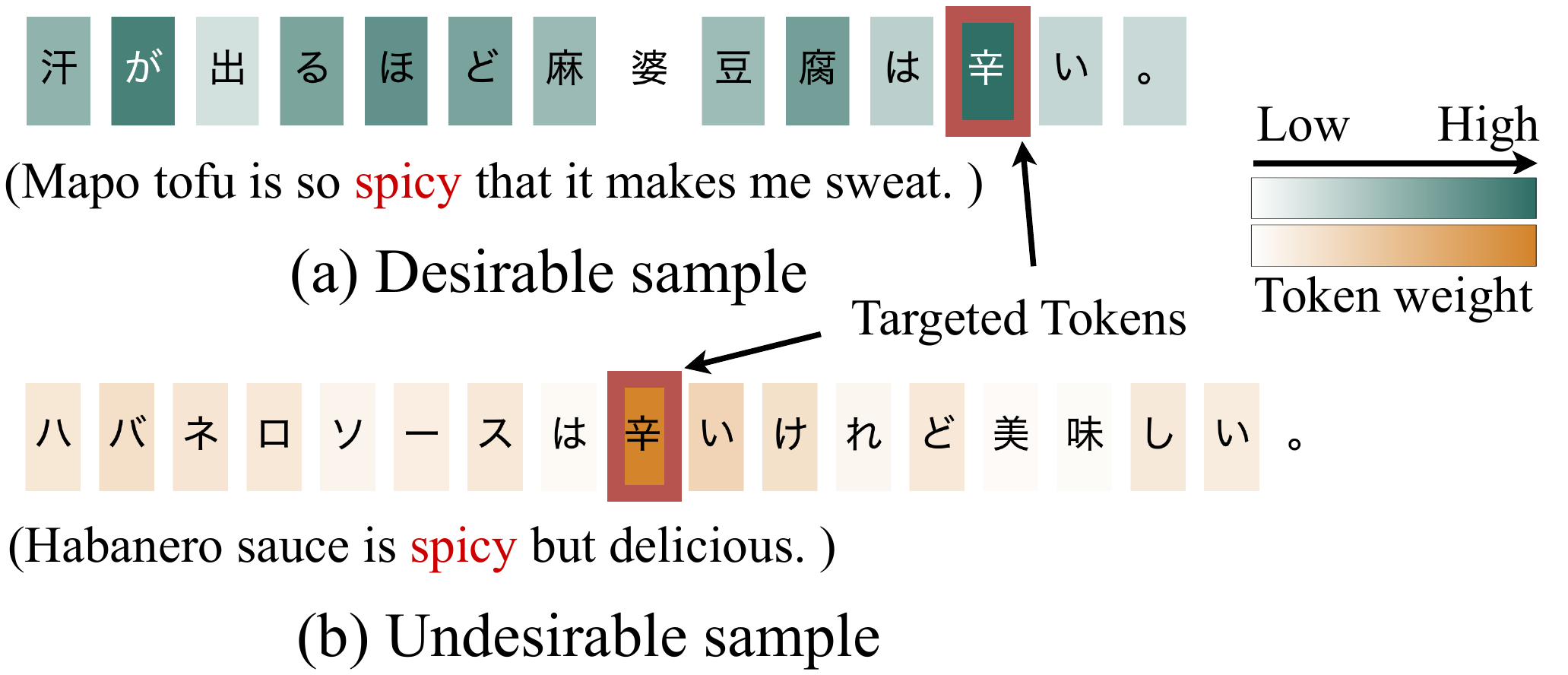}
  \caption{Case study of token weight estimation. The tokens of target character have higher weights.}
  \label{fig:weight}
\end{figure}

\paragraph{Case Study.}
Figure~\ref{fig:weight} presents a case study showing that the tokens corresponding to the target characters receive notably higher weights compared to non-targeted tokens.

\begin{figure}[t]
  \centering
    \centering
    \includegraphics[width=\linewidth]{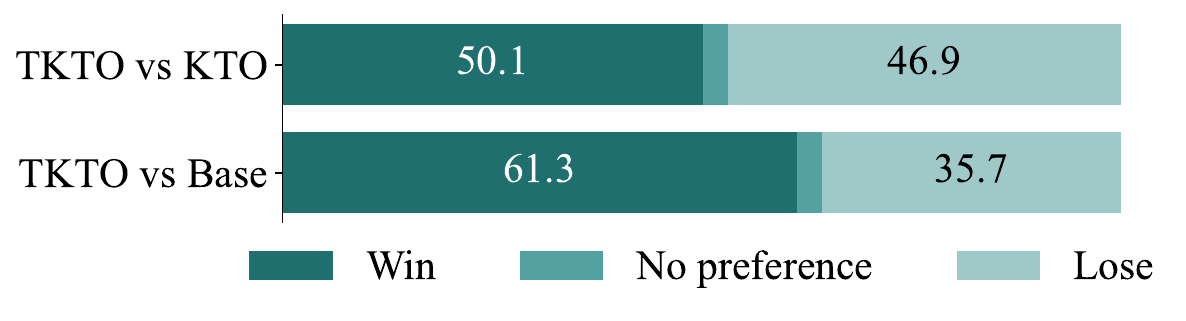}
  \caption{Results for ABX preference test.}
  \label{fig:abx}
\end{figure}

\begin{table}[t]
  \centering
  \resizebox{0.65\linewidth}{!}{%
  \begin{tabular}{lccc}
    \toprule
    & {Base} & {KTO} & {TKTO (ours)} \\
    \midrule
    \textbf{NMOS ↑} & 4.09 & 4.17 & \textbf{4.21} \\
    \bottomrule
  \end{tabular}
  }
  \caption{\diff{NMOS comparison (higher is better).}}
  \label{tab:nmos}
\end{table}

\subsection{Subjective Evaluation Results}
Table~\ref{tab:nmos} presents the NMOS subjective evaluation scores. Figure~\ref{fig:abx} shows the results of the ABX test.
In the subjective evaluations, TKTO also outperforms the base model and the standard KTO.

\subsection{Computational Cost and Data Efficiency}
\paragraph{Building Two Contrastive LLMs.}
Our TKTO framework requires training two additional contrastive LLMs to estimate token weights, which introduces an extra pre-computation step in the pipeline. However, the additional training cost is relatively small: in our setup, training these two contrastive LLMs takes approximately 10 minutes on 8×A100 GPUs. This overhead is negligible compared to the overall TTS training process, where pre-training on tens of thousands of hours of data typically requires weeks of computation.

\paragraph{Data Efficiency with Limited Pairs.}
In our dataset, only 10.5\% of utterances contain both a desirable and an undesirable output, while 89.5\% contain only one side. This imbalance occurs because modern TTS systems tend to produce consistently correct or incorrect pronunciations, making paired samples rare. As a result, paired methods such as DPO can use only 1.5K samples, whereas unpaired methods can utilize 9K samples. In our experiments, TKTO is able to leverage the unpaired samples, resulting in approximately 6× more usable data than what is accessible to paired preference optimization methods.

\subsection{\diff{Additional Validation on Chinese}}
\diff{
In addition to Japanese, we evaluate the generalization capability of our method on Chinese.
Following the same experimental setup as in the Japanese evaluation, we generate 5,000 sentences containing the polyphonic character 
\begin{CJK}{UTF8}{min} "行" \end{CJK}
which has two context-dependent readings, “xíng” and “háng.”
For each sentence, we generate five speech samples using the base TTS model with a Chinese speaker, resulting in 1.5K paired samples and 4K unpaired samples.
We evaluate the correctness of each reading using the wav2vec2-mms-1b-cmn-phonetic checkpoint\footnote{https://huggingface.co/Chuatury/wav2vec2-mms-1b-cmn-phonetic}.
\paragraph{Results.}
Table \ref{tab:chinese_validation} presents the Chinese evaluation results.
TKTO achieves the highest accuracy and the lowest CER and bad-case ratio among all compared methods.
These results demonstrate that TKTO generalizes beyond Japanese and is effective for polyphonic disambiguation in Chinese.
}

\begin{table}[t]
\centering
\resizebox{\linewidth}{!}{%
\begin{tabular}{l l c c c}
\toprule
Moethods & PO Data & Acc $\uparrow$ & CER $\downarrow$ & Bad $\downarrow$ \\
\midrule
Base model      & --        & 0.645 & 0.024 & 0.010 \\
DPO             & Paired    & 0.682 & 0.021 & 0.007 \\
KTO             & Paired    & 0.712 & 0.026 & 0.009 \\
KTO             & Unpaired  & 0.832 & 0.018 & 0.010 \\
\rowcolor{gray!15} TKTO (ours)     & Paired    & 0.746 & 0.024 & 0.007 \\
\rowcolor{gray!15} TKTO (ours)     & Unpaired  & \textbf{0.898} & \textbf{0.014} & \textbf{0.003} \\
\bottomrule
\end{tabular}
}
\caption{\diff{Results on Chinese polyphonic disambiguation.}}
\label{tab:chinese_validation}
\end{table}

\section{Conclusion}
\label{sec:conclusion}
We propose TKTO that eliminates the need for paired data, enabling a more data-efficient training paradigm, and directly targets token-level units, automatically providing fine-grained alignment signals without token-level annotations.
TKTO serves as an off-policy approach compatible with human feedback, while extending it to an on-policy setting remains a promising direction for future work.


\section*{Limitations}
\label{sec:limitation}
The evaluations are conducted on only challenging Japanese \diff{and Chinese} data.
Although multiple preference optimization methods and configurations were tested, 
we only evaluate a single model size (0.5B), as larger versions of the same backbone are not publicly available and computational constraints prevent extensive pretraining and scaling experiments. 
Investigating scaling behavior and extending the approach to other model architectures remain to be explored.
While this work only focuses on preference optimization for TTS, the proposed TKTO framework can potentially be applied to other text generation tasks where specific tokens play a crucial role.

\section*{Acknowledgments}
The authors would like to sincerely thank the anonymous reviewers for their valuable comments and helpful suggestions.

\section*{Ethical Considerations}
\label{sec:ethical}
While text-to-speech (TTS) technology may raise concerns regarding unauthorized generation or misuse, the models utilized in this paper were used for research purposes under controlled conditions.

\bibliography{custom}

\appendix

\section*{Appendix}
\label{sec:appendix}
\section{\diff{Kahneman--Tversky Optimization}}
\label{sec:kto}
\diff{
Kahneman--Tversky Optimization (KTO) \cite{kto} is a pair-free alignment objective that directly maximizes a proxy of human utility using binary feedback indicating whether an output is \emph{desirable} or \emph{undesirable}.  
Unlike DPO \cite{dpo}, KTO does not require paired comparisons.
}

\paragraph{\diff{Reference-adjusted reward.}}
\diff{
Given an input $x$, output $y$, current policy $\pi_\theta$, and reference policy $\pi_{\text{ref}}$, we define the reward $r_\theta(x,y)$ as:
\begin{equation}
r_\theta(x,y) = \log \frac{\pi_\theta(y\mid x)}{\pi_{\text{ref}}(y\mid x)} .
\end{equation}
A reference point $z_0$ is defined as:
\begin{equation}
z_0 = \mathrm{KL}\!\left(\pi_\theta(\cdot\mid x)\,\|\,\pi_{\text{ref}}(\cdot\mid x)\right),
\end{equation}
which captures the expected reward.
}

\paragraph{\diff{Value function.}}
\diff{
To model loss aversion and diminishing sensitivity, KTO uses a logistic value function:
\begin{equation}
v(x,y) =
\begin{cases}
\lambda_D\, \sigma\!\left(\beta (r_\theta(x,y)-z_0)\right), \\ \quad \quad \quad \text{if } y \sim y_\text{desirable}\mid x,\\
\lambda_U\, \sigma\!\left(\beta (z_0-r_\theta(x,y))\right), \\ \quad \quad \quad \text{if } y \sim y_\text{undesirable}\mid x,
\end{cases}
\end{equation}
where $\sigma(\cdot)$ is the sigmoid function, $\beta$ controls risk sensitivity, and
$\lambda_D,\lambda_U$ control gains and losses.
}

\paragraph{\diff{KTO objective.}}
\diff{
The KTO loss is defined as:
\begin{equation}
\mathcal{L}_{\text{KTO}}(\pi_\theta)
= \mathbb{E}_{(x,y)\sim\mathcal{D}}
\big[\, - v(x,y) \,\big],
\end{equation}
}

\paragraph{\diff{Connection to TKTO.}}
\diff{
Our TKTO extends KTO by lifting the optimization from the sequence level to the token level for TTS.
While KTO assigns a single utility to an entire output $y$, TKTO introduces token-wise weights that estimate the relative contribution of each token to human utility.
By explicitly modeling heterogeneous token importance, TKTO enables fine-grained credit assignment across tokens,
allowing the model to emphasize critical reasoning or decision-making steps while down-weighting less informative tokens.
In this sense, TKTO can be viewed as a token-wise generalization of KTO that preserves its prospect-theoretic inductive bias
while substantially improving optimization granularity.
}

\section{Implementation Details}
\label{app:implement}
Our implementation was based on the publicly available codes of prior work \cite{f5,cosyvoice2}, which are released under research-permissive licenses, and hyperparameters and libraries used followed those studies.
Base model and baseline F5-TTS were initialized from the pretrained checkpoints and fine-tuned once.
Training took a few minutes on 8$\times$A100 GPUs.
All preference optimization experiments were conducted on 1 epoch with 1e-6 learning rate.
We set $\lambda_D=\lambda_U=1$, $\beta=0.10$, we set $\mu=1$ for desirable tokens, $\mu=-1$ for undesirable ones, following \citet{kto}.

\section{Parameter Sensitivity Analysis}
\begin{table}[t]
  \centering
  \resizebox{\linewidth}{!}{%
  \begin{tabular}{l|ccc|ccc}
    \toprule
      Parameter & \multicolumn{3}{c|}{Female} & \multicolumn{3}{c}{Male} \\
                                                   L, U  & Acc ↑ & CER ↓ & Bad ↓ & Acc ↑ & CER ↓ & Bad ↓ \\
    \midrule
    -1, 1   & 0.937 & 0.076 & 0.028 & 0.951 & 0.088 & 0.031 \\
    -2, 2   & 0.949 & 0.075 & 0.029 & 0.958 & 0.085 & 0.027 \\
    -3, 3   & 0.956 & 0.072 & 0.027 & 0.948 & 0.088 & 0.028 \\
    \bottomrule
  \end{tabular}
  }
  \caption{Parameter sensitivity analysis.}
  \label{tab:sensitivity}
\end{table}

\begin{table}[t]
\centering
\small
\begin{tabular}{lcccc}
\toprule
& \multicolumn{2}{c}{Female} & \multicolumn{2}{c}{Male} \\
\cmidrule(lr){2-3} \cmidrule(lr){4-5}
Model & CER $\downarrow$ & Bad $\downarrow$ & CER $\downarrow$ & Bad $\downarrow$ \\
\midrule
Base model & 0.152 & 0.110 & 0.151 & 0.094 \\
KTO & 0.072 & 0.050 & 0.076 & 0.047 \\
TKTO (ours) & \textbf{0.065} & \textbf{0.044} & \textbf{0.069} & \textbf{0.045} \\
\bottomrule
\end{tabular}
\caption{Evaluation results using the \texttt{nvidia/parakeet-tdt\_ctc-0.6b-ja} ASR model.}
\label{tab:asr_robustness}
\end{table}

Since the clamping range controls the scaling of the reward weights, it plays an important role in balancing learning stability and sensitivity.
Table~\ref{tab:sensitivity} presents the clamping range (L,U) sensitivity analysis results.
While performance remains stable for moderate clamping ranges, wide bounds (e.g., (-3, 3)) can slightly lead to accuracy degradation.
This suggests that imposing reasonable constraints on the reward range is beneficial for achieving optimal results.
We select $(-2, 2)$ as the optimal range for experiments.

\section{Robustness across ASR Backends}
To verify the robustness of our evaluation, we conduct additional experiments using the \texttt{parakeet-tdt\_ctc-0.6b-ja}\footnote{\url{https://huggingface.co/nvidia/parakeet-tdt_ctc-0.6b-ja}} ASR model with paired preference data. Table~\ref{tab:asr_robustness} shows consistent improvements over the baselines, mirroring the trend observed with Whisper. As shown in Table~\ref{tab:asr_robustness}, TKTO achieves the lowest CER and Bad ratios for both speakers, confirming that the performance gains are not tied to a particular ASR backend.

\end{document}